\documentclass[conference]{IEEEtran}
\IEEEoverridecommandlockouts

\usepackage{cite}
\usepackage{amsmath,amssymb,amsfonts}
\usepackage{algorithmic}
\usepackage{graphicx}
\usepackage{hyperref}
\usepackage{multirow}
\usepackage{booktabs}
\usepackage{textcomp}
\usepackage{xcolor}
\def\BibTeX{{\rm B\kern-.05em{\sc i\kern-.025em b}\kern-.08em
    T\kern-.1667em\lower.7ex\hbox{E}\kern-.125emX}}
\begin{document}

\title{Dissecting the Dental–Lung Cancer Axis via Mendelian Randomization and Mediation Analysis\\
}
\author{
    Wenran Zhang*\textsuperscript{1}, Huihuan Luo*\textsuperscript{2}, Linda Wei\textsuperscript{3}, Nie Ping**\textsuperscript{1}, Yiqun Wu**\textsuperscript{4}, Dedong Yu**\textsuperscript{1}%
    \thanks{*:Wenran Zhang and Huihuan Luo are co-first authors}
    \thanks{**:Corresponding Authors}
    \thanks{\textsuperscript{1}Department of Second Dental Center, Shanghai Ninth People's Hospital, Shanghai Jiao Tong University School of Medicine, Shanghai, China. Emails: zhangwenran@sjtu.edu.cn,nieping@alumni.sjtu.edu.cn, yiqunwu@hotmail.com, yudedong@sjtu.edu.cn}%
    \thanks{\textsuperscript{2}School of Public Health, Key Lab of Public Health Safety of the Ministry of Education and NHC Key Lab of Health Technology Assessment, Fudan University, Shanghai, China. Email: hhluo21@m.fudan.edu.cn}%
    \thanks{\textsuperscript{3}Multimedia Laboratory, The Chinese University of Hong Kong, Hong Kong SAR, China. Email: 1155230127@link.cuhk.edu.hk}%
}

\maketitle

\begin{abstract}
Periodontitis and dental caries are common oral diseases affecting billions globally. While observational studies suggest links between these conditions and lung cancer, causality remains uncertain. This study used two-sample Mendelian randomization (MR) to explore causal relationships between dental traits (periodontitis, dental caries) and lung cancer subtypes, and to assess mediation by pulmonary function. Genetic instruments were derived from the largest available genome-wide association studies, including data from 487,823 dental caries and 506,594 periodontitis cases, as well as lung cancer data from the Transdisciplinary Research of Cancer in Lung consortium. Inverse-variance weighting was the main analytical method; lung function mediation was assessed using the delta method. The results showed a significant positive causal effect of dental caries on overall lung cancer and its subtypes. Specifically, a one-standard-deviation increase in dental caries incidence was associated with a 188.0\% higher risk of squamous cell lung carcinoma (OR = 2.880, 95\% CI = 1.236–6.713, p = 0.014), partially mediated by declines in forced vital capacity (FVC) and forced expiratory volume in one second (FEV1), accounting for 5.124\% and 5.890\% of the total effect. No causal effect was found for periodontitis. These findings highlight a causal role of dental caries in lung cancer risk and support integrating dental care and pulmonary function monitoring into cancer prevention strategies. The project will be open-sourced upon acceptance.

\end{abstract}


\section{Introduction}
Lung cancer is the second most common diagnosed cancer globally and a leading cause of cancer mortality \cite{1}. For a long time, lung cancer has been wrongly perceived as a "smoker's disease." However, as smoking rates decline, the proportion of never-smokers in lung cancer cases is increasing, accounting for approximately 10\% to 25\% \cite{2}. With this shift, researchers are dedicated to finding additional risk factors.

Oral disorders are considered to be associated with an increased risk of lung cancer. The respiratory tract is connected to the oral cavity, making the oral microbiome one of the main sources of the respiratory pathogens \cite{3,4,5}. Various respiratory pathogens have been detected in dental plaques, periodontal pockets and saliva, providing a basis for the subsequent invasion and infection by these pathogens in the lower respiratory tract \cite{6}. In parallel, epidemiological studies indicated that people with lower microbial diversity and richness of salivary microbiota have a higher risk of developing lung cancer, regardless of smoking status \cite{4,7,8}. Periodontitis and dental caries are two of the most prevalent microbially induced oral disorders. It can be hypothesized that there may exist a pathway between two dental traits and lung cancer without the bias of smoking. Several clinical studies have reported the association, however, the reliability of their results is questioned due to specific limitations \cite{9,10,11,12}. On one hand, clinical studies must adhere to strict ethical guidelines, which can restrict certain types of research designs or interventions. On the other hand, participant compliance, limited time and resources may restrict the sample size, study duration, or scope, potentially impacting the validity and reliability of the research. Therefore, further investigation using an unbiased and complementary approach is warranted to explore the potential causal relationship between oral lesions and lung cancer.

Mendelian randomization (MR) is an emerging approach for evaluating the causal effect of an exposure on a certain outcome, usually a disease, by utilizing single-nucleotide polymorphisms (SNPs) as instrumental variables (IVs) \cite{13}. In comparison to observational epidemiological studies, MR is less biased by reverse causation and confounding factors and has thus been widely used in aetiology analyses. Two-sample MR, as its name suggests, extracts genetic variants from different datasets, consequently resulting in more robust conclusions \cite{14}. However, no MR study has assessed the relationship between oral traits and lung cancer.

Dysbiosis of the oral microbiome promotes lung tumorigenesis through multiple mechanisms, including increased genetic toxicity and virulence effects, metabolic disruption, immune response, and pro-inflammatory actions \cite{15}. Probably due to similar pathways \cite{16,17,18}, observational studies proposed that impaired pulmonary function could also predict an increased risk of lung cancer \cite{16,17,18,19,20,21}. Another long-term cohort study showed that individuals with a greater degree of periodontitis had poorer lung function \cite{22}. Given that pulmonary disfunction may be related to both oral and lung lesions, we anticipate confirming these findings and exploring the potential mediating role of lung function in the pathway from oral to lung cancer. As our understanding of genetic and molecular pathways between dental traits and lung cancer deepens, it may contribute to future development of tumor prevention and screening. Therefore, in this study, we aimed to investigate the effect of periodontitis and dental caries on the risk of lung cancer, and to quantify the role of lung function as a mediator.
\section{Methods}\label{sec2}
\begin{figure*}[h]
    \centering
    \includegraphics[width=\textwidth]{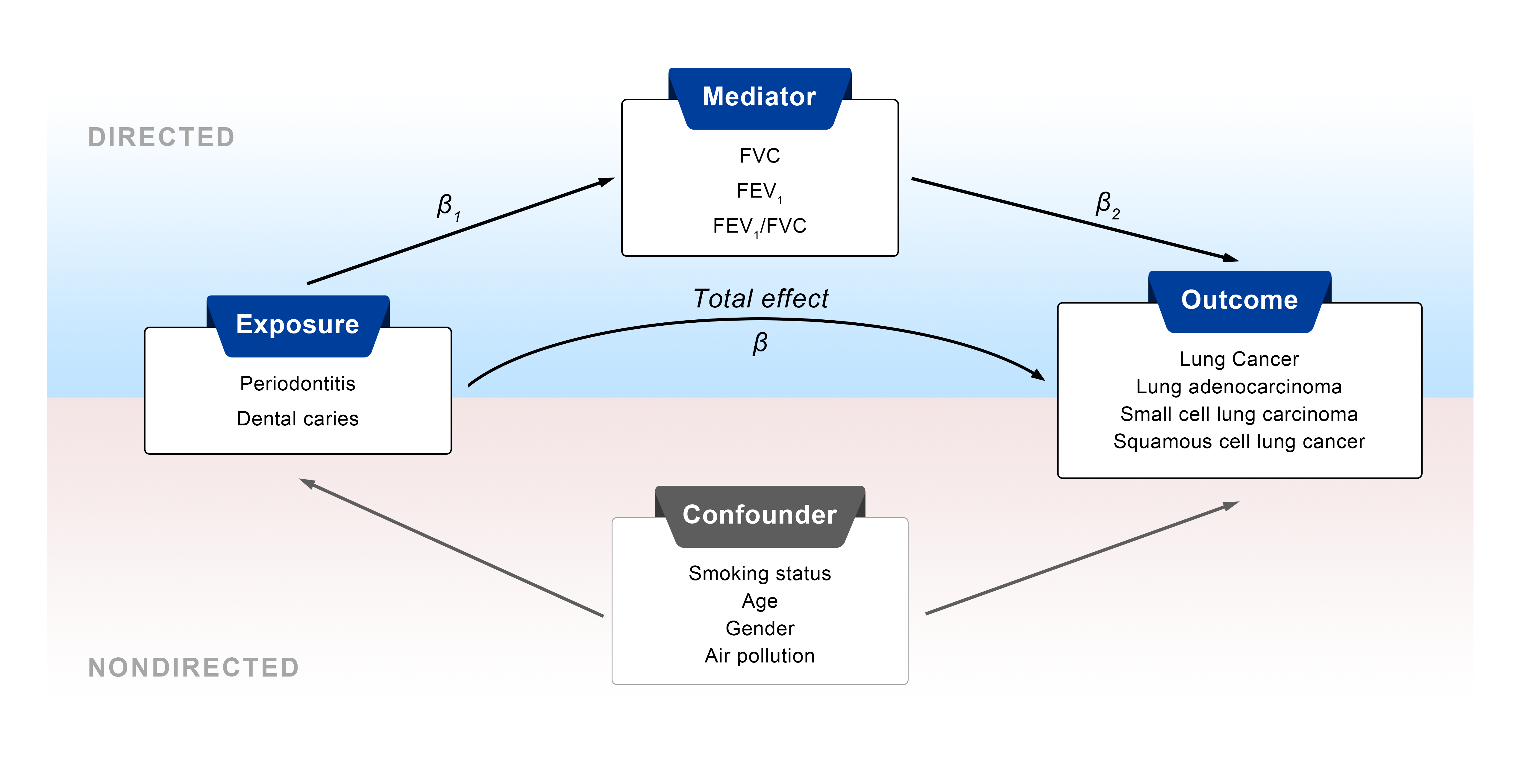}
    \caption{Directed Acyclic Graph (DAG) of the association between dental traits and lung cancer. The blue bar represents the directed effect of the exposure on the outcome and lung function impairment may play a mediating role. The red bar shows main confounders related to dental traits and lung cancer. Bias can be reduced by adjusting or controlling for these confounders. $\beta_1$: causal effect of oral traits on mediators; $\beta_2$: causal effect of mediators on lung cancer and its subtypes; $\beta$: total effect of oral traits on lung cancer and its subtypes.}
    \label{fig:fig1}
\end{figure*}

\subsection{Study design}
In this two-step, summary-based MR study, we explored the influences of dental diseases on different types of lung cancer. Moreover, the potential mediation of respiratory function parameters, including forced vital capacity (FVC), forced expiratory volume in 1 second ($\text{FEV}_1$) and $\text{FEV}_1/\text{FVC}$, between dental conditions and lung cancer was analysed. \autoref{fig:fig1} shows the analysis procedure of our study.

\subsection{Data sources and instrumental variable selections}
\subsubsection{Exposures: Periodontitis and dental caries}
The GWAS summary statistics of periodontitis and dental caries were obtained by Shungin et al. \cite{23}, who synthesized samples of European ancestry from the Gene–Lifestyle Interactions in Dental Endpoints (GLIDE) consortium and UK Biobank. It is the largest and most recent dataset for the time being. The genetic determinants of periodontitis and dental caries were separately identified from GLIDE (n = 44,563 and n = 26,792). The GWAS was adjusted for age, $\text{age}^2$ and other study-specific covariates. To ensure that the genetic variants strongly correlated with exposure, we set the P value threshold to $5\times 10^{-8}$ for dental caries and broadened it to $5\times 10^{-6}$ for periodontitis to obtain more SNPs \cite{24,25,26}. Those with limited linkage disequilibrium ($r^2 < 0.001$ and clump window > 10,000 kb) at the same time were employed as IVs. To further verify the genetic correlation, we calculated the F-statistic, which should be greater than 10 \cite{27}.

\subsubsection{Outcome: Lung cancer}
GWAS summary statistics for lung cancer and its subtypes (including lung adenocarcinoma, small cell lung carcinoma and squamous cell lung cancer) were obtained from the Transdisciplinary Research of Cancer in Lung of the International Lung Cancer Consortium (TRICL-ILCCO); 29,266 cases and 56,450 controls of European ancestry were included \cite{28}. The GWAS was adjusted for sex, age, and principal components. The basic information of the GWASs is shown in \autoref{table1}.

\begin{table}
\centering
\caption{Summary of the GWASs included in our study.}
\label{table1}
\resizebox{0.48\textwidth}{!}{%
\begin{tabular}{@{}ccccc@{}}
\toprule
Role & Traits & GWAS ID & Consortium & Sample sizes \\ \midrule
\multirow{2}{*}{Exposure} & Periodontitis & Not Applicable & GLIDE & 44563\\
 & Dental Caries & Not Applicable & GLIDE & 26792 \\ \addlinespace
\multirow{3}{*}{Mediator} & FVC & ebi-a-GCST007429 & SpiroMeta & 400102 \\
 & FEV\textsubscript{1} & ebi-a-GCST007432 & SpiroMeta & 400102 \\
 & FEV\textsubscript{1}/FVC & ebi-a-GCST007431 & SpiroMeta & 400102 \\ \addlinespace
\multirow{4}{*}{Outcome} & Lung Cancer & ieu-a-987 & TRICL& 85716 \\
 & Lung adenocarcinoma & ieu-a-984 & TRICL &65864 \\
 & Small cell lung carcinoma & ieu-a-988 & TRICL &23371\\
 & Squamous cell lung cancer & ieu-a-989 & TRICL &62467 \\
\bottomrule
\end{tabular}
}
\end{table}

\subsubsection{Mediators: FVC, $\text{FEV}_1$ and $\textbf{FEV}_1/\text{FVC}$}
The GWAS summary statistics of lung function were integrated by Shrine et al. \cite{29}, who analysed 400,102 individuals of European ancestry (321,407 from UKB and 79,055 from SpiroMeta). The GWAS was adjusted for age, $\text{age}^2$, sex, height, and smoking status. SNPs significantly associated with impaired lung function at the genome-wide level (P value < $5\times 10^{-8}$) and without high linkage disequilibrium ($r^2 < 0.001$ and clump window > 10,000 kb) were selected. Similarly, we computed the F-statistic to verify the strength of the relationships.

\subsection{Statistical analysis}
\subsubsection{Primary MR analysis}
We first compared IVs of periodontitis and caries to confirm if any common gene variants existed. The inverse-variance weighting (IVW) method was used as the major analysis method for assessing the estimates \cite{30}. To further eliminate interference, we searched the PhenoScanner database for the above-mentioned instruments for potential confounders, including obesity, smoking status, alcohol consumption, etc., and deleted them \cite{31}. Then, MR analysis was conducted after the deletion, which served as the master model. However, instrument variants may cause horizontal pleiotropy, which means that other factors influence the outcome through causal pathways in addition to our exposure, thereby violating MR assumptions. Accordingly, the results from MR-Egger regression and the weighted median method can play an auxiliary role and are known to be relatively robust to horizontal pleiotropy \cite{32}. The weighted median method can provide consistent causal estimates when up to 50\% of the selected genetic instruments are invalid \cite{33}, and the MR-Egger method could detect and adjust for directional pleiotropy \cite{34}. Finally, we used the non-centrality parameter (NCP) to calculate the statistical power of our results \cite{35}.
\subsubsection{Sensitivity analysis}
To assess pleiotropy accurately, we conducted MR-Egger regression and the MR Pleiotropy RESidual Sum and Outlier (MR-PRESSO) test, which can identify additional outliers through the global and SNP-specific observed residual sum of squares, determining whether there is an evident difference after the removal of outliers \cite{36}. Cochran’s Q test was subsequently conducted to assess the heterogeneity of causal effect estimates between oral traits and lung cancer incidence.

Given that previous research has suggested that height could bias the MR analysis of pulmonary function \cite{37}, we additionally adjusted for height from different GWASs using multivariable MR (MVMR) when FVC, $\text{FEV}_1$ and $\text{FEV}_1/\text{FVC}$ were used as exposures \cite{38}. Compared with conventional methods in which instruments associated with height are immediately removed, MVMR does not weaken statistical power \cite{39}.
\subsubsection{MR mediation analysis}
To determine whether pulmonary function mediates the statistically significant association between exposure and outcome, we performed a mediation MR analysis by applying the two-step MR approach. In the first step, we calculated the causal effect of exposure on each mediator ($\beta_1$); in the second step, we estimated the causal effect of the mediators on lung cancer and its main subtypes ($\beta_2$). For combinations exhibiting significance (P value < 0.05) in both steps, $\beta_1*\beta_2$ represents the indirect effect of the mediator on the mechanism. $\beta$ represents the total effect of the exposures on the outcomes. Accordingly, we used the delta method to calculate the mediation proportion of lung function (mediation proportion = $\frac{\beta_1*\beta_2}{\beta}$) \cite{40}. To ensure the robustness of the mediation analysis, 95\% confidence intervals (CIs) were estimated by the bootstrap method. A confidence interval of zero indicates a nonsignificant mediating effect at the significance level of 5\% \cite{41}.

We used the following packages in R (version 4.3.1) for the analyses: TwoSampleMR (version 0.5.7) to conduct MR analyses; MRPRESSO (version 1.0), MRInstruments (version 0.3.2) and MRPracticals (version 0.0.1) to conduct sensitivity analysis and confirm outliers; and RMediation (version 1.2.2) to calculate the mediation.
\section{Results}\label{sec3}

\subsection{Characteristics of the selected SNPs}
After excluding SNPs that did not follow the above criteria, 22 SNPs of periodontitis and 20 SNPs of dental caries were selected as IVs to evaluate the effect of the exposure. And there was no genetic overlap between two groups. For lung function, 241 (FVC), 271 ($\text{FEV}_1$) and 305 ($\text{FEV}_1/\text{FVC}$) SNPs were selected as IVs to evaluate the effect of the mediators. The F-statistics of all those IVs were > 10, indicating no evidence of weak instrument bias.
\subsection{Effect of periodontitis and dental caries on lung cancer and its subtypes}
For periodontitis, after the harmonization of exposure and outcome, the incompatible allele (rs1807019) was deleted. The initial MR results indicated no significance (IVW: OR = 1.172, 95\% CI = 0.833-1.650, p = 0.362). Then, we searched the PhenoScanner database, which indicated that rs62177307 was significantly associated with height traits and was therefore removed. 20 IVs were determined as the final instrumental variables referring to periodontitis. Using univariate MR, we found that periodontitis had no significant effect on lung cancer incidence (IVW: OR = 1.137, 95\% CI = 0.797-1.623, p = 0.479). We identified one outlier (rs12568187) through the MR-PRESSO Outlier test, and the P value in the MR-PRESSO distortion test was 0.034, which indicated that the outlier had an evident effect on the results. However, there was still no significance after its elimination (IVW: OR = 0.546, 95\% CI = 0.259-1.151, p = 0.130).

\begin{figure*}[h]
    \centering
    \includegraphics[width=\textwidth]{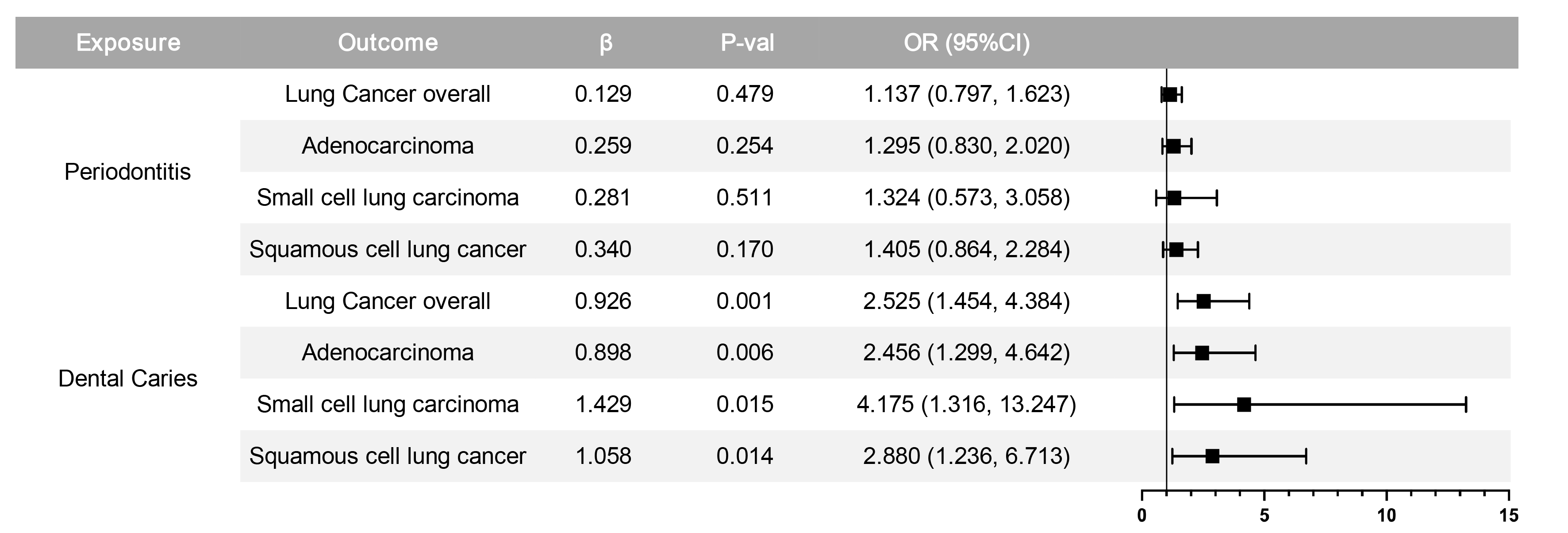}
    \caption{MR results (IVW) and forest plots summarizing the causality of dental traits on lung cancers.}
    \label{fig2}
\end{figure*}

Similarly, for dental caries, the palindromic SNP (rs3865314) was deleted. No confounding SNP was found in the PhenoScanner database. Those confounders were deleted, and 15 SNPs were subsequently selected as the final IVs. We found that dental caries was positively associated with a higher risk of lung cancer (IVW: OR = 2.525, 95\% CI = 1.454-4.384; p = 0.001). While the results from Cochran’s Q test confirmed the presence of heterogeneity (IVW: Q=30.812, p=0.006), the MR-Egger intercept did not indicate evidence of pleiotropy (intercept=-0.003, p=0.915) (shown in \autoref{fig2}). We used the MR-PRESSO Outlier test and identified one outlier (rs80270335); however, the P value of the MR-PRESSO distortion test (p = 0.555) showed that eliminating the outlier did not significantly affect the results.

Furthermore, the genetic correlations between oral traits and lung cancer subtypes are presented in \autoref{fig2}. Dental caries was positively associated with adenocarcinoma (IVW: OR = 2.456, 95\% CI = 1.299-4.642, p = 0.006), small cell lung carcinoma (IVW: OR = 4.175, 95\% CI = 1.316-13.247, p = 0.015) and squamous cell lung cancer (IVW: OR = 2.880, 95\% CI = 1.236-6.713, p = 0.014), as well as overall lung cancer. And all of them were proved to possess adequate statistical power (approximate to 1.00). However, periodontitis had no significant influence on any of the lung cancer subgroups.

To strengthen causal reasoning, we also employed the MR-Egger and the weighted median to evaluate the impacts, and the results showed relatively high robustness. According to the sensitivity analysis, we used MR-Egger regression and found that dental caries showed no pleiotropy with any of the subgroups. The MR-PRESSO outlier test was used to detect and exclude outliers in our analysis, and the outcomes after elimination, if the results were positive, were in accordance with the initial results.
\subsection{Effect of dental caries on lung function}
Given that periodontitis had no impact on the outcome, we focused our analysis solely on examining the mediating effects between dental caries and the incidence of lung cancer. After the harmonization of exposure and mediators, incompatible alleles, palindromic SNPs, alleles unable to be located in mediators and confounders searched in the PhenoScanner were deleted. The results of the IVW method revealed that the occurrence of dental caries can lead to lower values of FVC (IVW: OR = 0.819, 95\% CI = 0.726-0.923, p = 0.001) and $\text{FEV}_1$ (IVW: OR = 0.857, 95\% CI = 0.769-0.954, p = 0.005). The results of the sensitivity analysis are shown in \autoref{table2}. MR-PRESSO distortion tests were applied to screen for outliers, the influence of which was not apparent in the results.
\begin{table}[ht]
\centering
\caption{MR results for the relationship between dental caries and lung function.}
\resizebox{0.48\textwidth}{!}{%
\begin{tabular}{lcccccc}
\toprule
\multirow{2}{*}{Outcome} & \multicolumn{2}{c}{IVW MR analysis} & \multicolumn{2}{c}{Heterogeneity test} & \multicolumn{2}{c}{Pleiotropy test} \\
\cmidrule(r){2-3} \cmidrule(r){4-5} \cmidrule(r){6-7}
& OR (95\% CI) & $P$ value & $Q$ & $P$ value & Intercept & $P$ value \\
\midrule
FVC & $0.819(0.726,0.923)$ & 0.001 & 28.766 & 0.004 & 0.010 & 0.043 \\
$\mathrm{FEV}_1 / \mathrm{FVC}$ & $1.024(0.875,1.197)$ & 0.771 & 63.467 & $2.868 \times 10^{-8}$ & -0.015 & 0.025 \\
$\mathrm{FEV}_1$ & $0.857(0.769,0.954)$ & 0.005 & 31.652 & 0.004 & 0.002 & 0.710 \\
\bottomrule
\end{tabular}%
}
\label{table2}
\end{table}
 
\subsection{Effect of lung function on lung cancer}
\begin{figure*}[h]
    \centering
    \includegraphics[width=\textwidth]{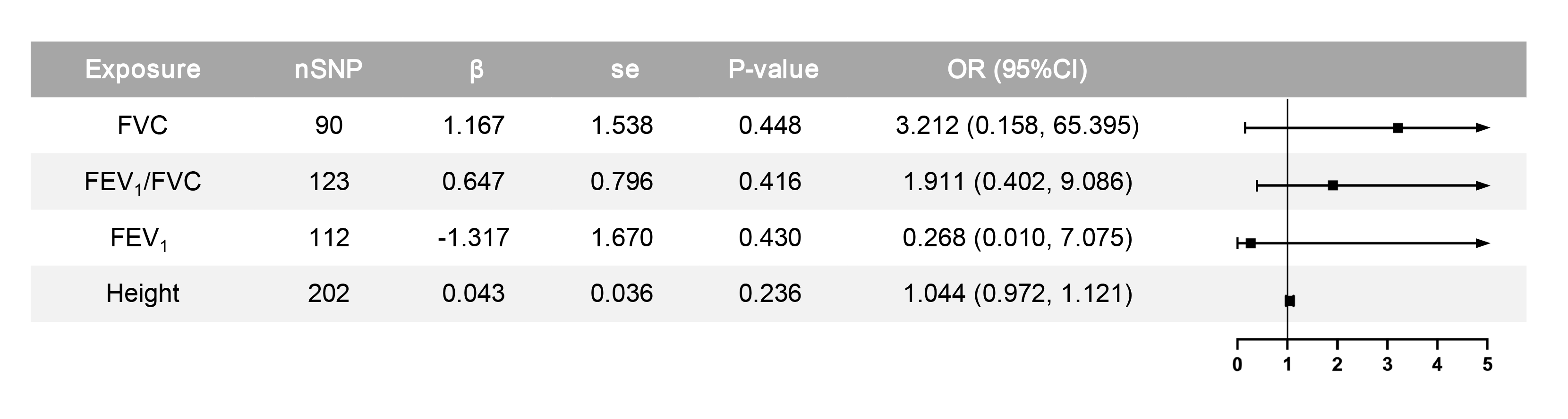}
    \caption{Results of multivariable MR analysis adjusted for the interference of height.}
    \label{fig3}
\end{figure*}

After the harmonization of the mediator and outcome data, incompatible alleles, palindromic SNPs and alleles unable to be located in the outcome were deleted. A total of 207 FVC, 240 $\text{FEV}_1$ and 276 $\text{FEV}_1/\text{FVC}$ SNPs were selected as the final IVs for parameters of lung function. All MR analyses suggest that indexes of lung function are not significantly associated with overall lung cancer. Furthermore, heterogeneity was observed between pulmonary function and lung cancer incidence. Pleiotropy was not found by MR-Egger regression analysis but did by the MR-PRESSO global test. Outliers were confirmed but played a negligible role in the results. Additional adjustment for height through MVMR did not change the conclusion, as shown in \autoref{fig3}. 

Further MR analyses of the subtypes of lung cancer are shown in \autoref{table3}. FVC was negatively correlated with squamous cell lung cancer ($\beta= -0.271$, p = 0.022), and a similar pattern of causality was observed for $\text{FEV}_1$ ($\beta= -0.402$, p = $6.38\times 10^{-5}$). In other words, impaired FVC and $\text{FEV}_1$ may increase the risk of squamous cell carcinoma. Sensitivity analyses of both mediations presented robust heterogeneity and no pleiotropy with lung cancer or its subtypes. MR-PRESSO tests were applied to confirm outliers; however, the elimination of these outliers did not make an evident difference (MR-PRESSO distortion test: p > 0.05).

\begin{table}[h]
\caption{MR analysis outcomes between lung function and lung cancer patients.}
\centering
\resizebox{0.48\textwidth}{!}{%
\begin{tabular}{@{}lccccc@{}}
\toprule
Exposure & Outcome & $\boldsymbol{\beta}$ & OR (95\% CI) & P value \\
\midrule
\multirow{4}{*}{FVC} & Lung cancer overall & -0.115 & $0.892(0.742,1.072)$ & 0.222 \\
& Adenocarcinoma & 0.053 & $1.054(0.865,1.284)$ & 0.601 \\
& Small cell lung carcinoma & -0.230 & $0.794(0.589,1.072)$ & 0.132 \\
& Squamous cell lung cancer & -0.271 & $0.762(0.604,0.962)$ & 0.022 \\
\addlinespace
\multirow{4}{*}{$\text{FEV}_1$} & Lung cancer overall & -0.149 & $0.861(0.738,1.005)$ & 0.058 \\
& Adenocarcinoma & -0.064 & $0.938(0.789,1.115)$ & 0.470 \\
& Small cell lung carcinoma & -0.151 & $0.860(0.666,1.110)$ & 0.248 \\
& Squamous cell lung cancer & -0.402 & $0.669(0.549,0.815)$ & $6.38 \times 10^{-5}$ \\
\bottomrule
\end{tabular}%
}
\label{table3}
\end{table}


\subsection{Mediation analysis}
Based on the results of MR analyses, we found that FVC and $\text{FEV}_1$ played mediating roles in the causal effect of dental caries on squamous cell lung cancer. We first calculated their indirect effects and divided them by the total effect of the exposure on the outcome. The proportions of mediation were 5.124\% and 5.890\%, respectively, and showed significance after the bootstrapping procedure, as shown in \autoref{tab:table4}.
\begin{table}[h]
\centering
\caption{Estimated mediation proportion of the relationship between dental caries and squamous cell lung cancer.}
\label{tab:table4}
\resizebox{0.48\textwidth}{!}{%
\begin{tabular}{cccccc}
\toprule
\multirow{2}{*}{Mediator} & \multirow{2}{*}{ $\beta$} & \multirow{2}{*}{$\beta_1$} & \multirow{2}{*}{$\beta_2$} & Mediation effect size & \multirow{2}{*}{Mediation proportion (\%)} \\
 &  &  &  & $\beta_1 * \beta_2$ (95\% CI) &  \\
\midrule
FVC & 1.058 & -0.200 & -0.271 & $0.054(0.006,0.121)$ & 5.124 \\
$\mathrm{FEV}_1$ & 1.058 & -0.155 & -0.402 & $0.062(0.016,0.123)$ & 5.890 \\
\bottomrule
\end{tabular}%
}

\end{table}

\section{Discussion}
We conducted a MR study to explore the causal relationship between dental traits and lung cancer incidence and found that a one-standard-deviation increase in caries incidence was associated with a 152.5\% increased lung cancer risk (IVW: OR = 2.525, 95\% CI = 1.454-4.384; p = 0.001). Similar relationships were also observed within the subgroups. However, there was no obvious evidence to support a causal relationship between periodontitis and lung cancer(IVW: OR = 0.546, 95\% CI = 0.259-1.151, p = 0.130). In addition, we observed that the causality between dental caries and squamous cell lung carcinoma is partially mediated by FVC and $\text{FEV}_1$, with mediation proportions accounting for 5.124\% and 5.890\%, respectively.

 It is generally assumed that both dental caries and periodontitis usually arise from an abnormal composition of dental plaque \cite{42}. Such dysbiosis makes the oral cavity a reservoir of respiratory pathogens \cite{43}. Consequently, respiratory diseases, including pneumonia, asthma, COPD and lung cancer, may occur \cite{4,43}. However, there are still several vital differences between both lesions, which probably lead to different MR results. 
 
First, their pathogens differ. \textit{Streptococcus mutans} and \textit{Lactobacilli} are recognized as caries-related pathogens, while \textit{Porphyromonas gingivalis}, \textit{Tannerella forsythia} and \textit{Treponema denticola} are identified as a pathogenic triad acting on periodontitis \cite{44}. After analysing the saliva of nonsmoking lung cancer patients, a greater abundance of the \textit{Bacilli} class and \textit{Lactobacillales} order was found associated with an increased risk of lung cancer, where the above caries-related pathogens were included. Meanwhile, a greater abundance of \textit{Spirochaetia} and \textit{Bacteroidetes} was associated with a decreased risk, where the periodontal pathogens were included \cite{4}. The opposite relevance can explain different MR results in our study. Second, the anatomical location where microbes colonize may play a part. For dental caries, plaques are usually situated on the surface of teeth and above the gingival margin and can migrate with saliva and airflow to a deeper anatomical position \cite{44}. In contrast, dental plaques for periodontitis are difficult to transfer because of the narrow, deep, hard-to-reach location of the periodontal pocket and gingival sulcus. Third, their pathogenic mechanisms are distinct from each other. Periodontitis is primarily caused by the interaction between bacteria and the host's immune response, while dental caries is caused by the acidic byproducts of bacterial metabolism that demineralize the tooth enamel \cite{45,46}. It explains why IVs of these two traits are completely inconsistent, which is in tune with the previous study \cite{47}.

In addition, our mediation analysis suggested that only a small part of the association between dental caries and squamous cell lung cancer was mediated by lung function (5.124\% for FVC and 5.890\% for $\text{FEV}_1$). One possible explanation for the lower proportion than expected might be that $\text{FEV}_1$, FVC and their ratio are only a rough proxy for dysfunction, which cannot completely replace the assessment of lung impairment. Moreover, in the relationship between oral conditions and lung cancer, the immune reactions, toxic effects, and inflammatory responses caused by abnormal oral microbiota can be a complex pathological process \cite{15}. This process may involve multiple mediating and moderating factors, with lung function being just one of them. 

What’s more, we took multiple measures to minimize bias caused by the smoking status and other possible confounders. We first calculated the F-statistic of IVs and eliminate all weak instruments to avoid the bias. Then we retrieved these IVs in the PhenoScanner database and deleted ones significantly correlated to other factors, especially smoking status. In sensitivity analysis, we conducted MR-PRESSO test to identified outliers which may cause pleiotropy and invalidate the significant results. However, in theory the additional confounding caused by collider bias could not be avoided completely \cite{48}. Moreover, the extent to which tobacco use interferes with periodontitis and dental caries may vary. There is a dose–response relationship between smoking status and risk of periodontitis, while the causality between tobacco use and caries has not been proved \cite{49,50}. It means that MR results of periodontitis are more likely to be interfered with by tobacco than those of caries. 

Our study presents new possibilities for lung cancer screening criteria, which mainly target smokers and elderly individuals for the time being. In a large-scale study on low-dose CT scanning in high-risk individuals, only 0.03\% of never-smokers were enrolled in the screening group \cite{51}. Based on our results, dental traits may be included into innovative screening guidelines. Furthermore, our findings may lead to positive progress in public health and clinical practice. Given the accumulating evidence that oral lesions are causative factors for respiratory disorders, effective therapy for individuals with oral diseases, especially those with caries and periodontitis, may be advantageous for protecting the respiratory system. A 2-year pilot trial provided a similar conclusion that periodontal therapy could improve the lung function of COPD patients \cite{52}. With additional supportive evidence emerging, regular oral health care may become one of the precautions taken against pulmonary lesions. Additionally, an in-depth understanding of the interaction mechanism between the oral environment and tumours will facilitate the design of subsequent new targeted drugs.

The strengths of our study can be summarized as follows. On one hand, this was the first two-sample MR study to identify the causal effect of oral lesions on various types of lung cancer. Numerous previous clinical trials have tried to explore the association between oral traits and pulmonary cancerization but are limited by restricted sample size, untenable follow-up, the influence of confounding factors, and most importantly, the possibility of upside-down causality. Therefore, our study mitigated such indeterminacy to some extent and opened up a novel possibility of the genetic pathway involved in lung cancer. On the other hand, from a statistical perspective, our summary data for dental caries and periodontitis were obtained from the largest GWAS datasets available, and summary data of exposure, mediators and outcomes were extracted from different GWAS consortiums, which could strengthen the statistical power \cite{53}. We also applied different MR sensitivity analyses and screened multiple confounders to minimize bias from horizontal pleiotropy or other sources.

Despite more robust conclusions obtained from our massive genetic study, there are still some limitations to be discussed. First, participants from four consortia were all of European ancestry to avoid ethnic deviation, but an unknown of other races followed, such as African and Asian populations. Second, the phenotypes of cancer were binary, so we could not assess potential dose-dependent changes \cite{54}. Third, lung cancer can be divided into different types, including lung adenocarcinoma, lung squamous cell carcinoma, small-cell lung cancer and large-cell lung cancer, which differ in their pathogenic mechanisms \cite{55}. The lack of subgroup dissection at the molecular level and the mere application of overall consideration may cause deviation. Fourth, neither GWAS nor observational studies can avoid bias caused by selecting survivors of lung cancer \cite{56}. Fifth, similarly, both methods cannot totally avoid influences of confounders such as smoking status \cite{48}.
\section{Conclusion}
Our two-step MR study detected sufficient genetic evidence that dental caries might increase the risk of pulmonary insufficiency and lung cancer. Additionally, we observed a mediating effect of FVC and $\text{FEV}_1$ between dental caries and squamous cell lung cancer, but we were unable to provide credible evidence to support the argument that periodontitis plays a role in the respiratory system, which has been confirmed by numerous clinical trials and meta-analyses. In view of the complexity of those pathways, the current findings should be treated with caution, and future studies should be conducted to explore the mechanism in diverse dimensions and to underpin these associations.

\bibliographystyle{IEEEtran}
\bibliography{IEEEabrv,mybibfile}

\end{document}